\documentclass[journal]{IEEEtran}

\usepackage{amsmath}
\usepackage{graphicx}
\usepackage[colorlinks=true, allcolors=blue]{hyperref}
\usepackage{caption}
\usepackage[nocompress]{cite}
\usepackage{multirow}
\usepackage{tabularx}
\usepackage{subcaption}
\usepackage{bm}
\usepackage{hyperref}
\usepackage{varwidth}
\usepackage{caption}
\usepackage{afterpage}
\usepackage{float}
\usepackage{xcolor}
\usepackage{soul}

\begin{document}

\title{VIO-DualProNet: Visual-Inertial Odometry with Learning Based Process Noise Covariance}

\author{Dan~Solodar and~Itzik~Klein

\thanks{The authors are with the Hatter Department of Marine Technologies, Charney School of Marine Sciences, University of Haifa, Israel.\\ E-mails: \{dsolodar@campus, kitzik@univ\}.haifa.ac.il}}

\maketitle
\begin{abstract}
Visual-inertial odometry (VIO) is a vital technique used in robotics, augmented reality, and autonomous vehicles. It combines visual and inertial measurements to accurately estimate position and orientation. Existing VIO methods assume a fixed noise covariance for the inertial uncertainty. However, accurately determining in real-time the noise variance of the inertial sensors presents a significant challenge as the uncertainty changes throughout the operation leading to suboptimal performance and reduced accuracy. To circumvent this, we propose VIO-DualProNet, a novel approach that utilizes deep learning methods to dynamically estimate the inertial noise uncertainty in real-time. By designing and training a deep neural network to predict inertial noise uncertainty using only inertial sensor measurements, and integrating it into the VINS-Mono algorithm, we demonstrate a substantial improvement in accuracy and robustness, enhancing VIO performance and potentially benefiting other VIO-based systems for precise localization and mapping across diverse conditions.
\end{abstract}
\section{Introduction}
Accurate estimation of a robot's six degrees of freedom (6DOF) motion is a crucial requirement in diverse fields including robotics, autonomous driving, drones, and augmented reality. It presents a significant challenge that requires robust and precise solutions to enable advancements in these domains. In recent years, many simultaneous localization and mapping (SLAM) techniques were developed, leveraging a variety of sensors, including stereo cameras 
\cite{engel2017direct}, \cite{newcombe2011dtam},  \cite{mur2017orb}, LiDAR \cite{hess2016real,khan2021comparative,zhang2014loam}, depth cameras \cite{hu2012robust, hai20153d}, ultra white-band sensors \cite{guo2016ultra,queralta2020uwb}, and magnetic sensors \cite{tan2020flydar}. Among these sensor combinations, Visual-Inertial Odometry(VIO)~\cite{scaramuzza2011visual, qin2018vins}
stands out as a preferred approach for many applications due to its light weight and low power consumption. VIO algorithms fuse visual and inertial measurements and can achieve high accuracy performance via optimization. Visual sensors provide detailed information about the environment through image sequences, allowing for feature tracking, scene reconstruction, and visual motion estimation. Inertial sensors, on the other hand, provide measurements of specific force and angular velocity that can be integrated over time, helping to mitigate scale ambiguity in monocular cameras and compensate for visual drift in challenging scenes.\\
The main challenge of VIO methods, especially when using low-cost sensors, is that both visual and inertial sensors suffer from noises and biases that lead to drift in position estimation. To deal with that, optimization-based methods use noise covariance matrices to imply how noisy each sensor's data is and take it into account during the optimization process. Current state-of-the-art VIO algorithms assume a constant noise covariance. However, the inertial noise variance (uncertainty)  changes in real-world applications due to the motion type, sensor health, weather conditions,  electrical disturbances and other environmental conditions.  Therefore, there is a need for an adaptive inertial noise covariance estimation method for improving the robustness of the VIO algorithms.\\
In recent years, there has been a growing interest in leveraging deep learning techniques for navigation, SLAM and VIO methods. Deep learning methods have shown promising results in improving feature matching, loop closure detection,  pose estimation, and other related navigation tasks \cite{cohen2023inertial, klein2022data}. Similarly, in VIO, deep learning can be utilized to learn complex relationships between sensor inputs and estimate key parameters for improving the overall system performance \cite{han2019deepvio, aslan2022hvionet}.\\ 
Inspired by the success of deep-learning approaches in navigation, we propose VIO-DualProNet, a visual-inertial odometry algorithm with online learning of the inertial process noise covariance. In this manner, our algorithm can cope with the variations in inertial uncertainty and provide an accurate and robust VIO approach. \\ 
The main contributions of this work are as follows:
\begin{enumerate}
\item \textbf{DualProNet}: A deep-learning network capable of online regressing the accelerometer and gyroscope measurement uncertainty using only the inertial readings. DualProNet can be applied to any optimization or estimation problems involving inertial sensors.
\item \textbf{VIO-DualProNet}: An adaptive, deep-learning based, inertial noise parameter tuning algorithm, dedicated to optimization-based, factor graph, VIO algorithms. 
\end{enumerate}
We provide a detailed derivation of VIO-DualProNet and its associated network architecture designed to regress the inertial noise uncertainty. We demonstrate that our approach outperforms current model-based VIO algorithms and provides an accurate navigation solution. 
\newline The rest of the paper is organized as follows: Section \ref{sec:RW} reviews related work in VIO algorithms, adaptive noise estimation, and deep learning for SLAM. Section \ref{sec:PA} presents our proposed approach in detail, including the architecture of the deep neural network and its integration with VINS-Mono. Section \ref{sec:MTF} presents the 
deep learning model training and fitting. Section \ref{sec:ER} gives an in-depth presentation and analysis of the experimental results and performance evaluation, comparing our method against the constant noise covariance approach. Finally, Section \ref{sec:CONC} concludes the paper and discusses future research directions.
\section{Related Work}\label{sec:RW}

\subsection{Visual-Inertial Odometry (VIO)}
Visual-Inertial Odometry (VIO) is a powerful technique that fuses visual and inertial measurements to estimate the camera's position and orientation. It has gained significant attention due to its applicability in various domains, such as robotics, drones, augmented reality, and autonomous vehicles. \\
VIO algorithms can be broadly categorized into filter-based and optimization-based approaches.
\begin{itemize}
    \item Filter-based methods, such as the extended Kalman filter (EKF) \cite{li2013high,barrau2015ekf}, utilize recursive state estimation to propagate the state forward in time, while incorporating new measurements to correct and update the state. The EKF and its variants have been widely used in VIO and SLAM due to their real-time performance. However, they have limitations in handling non-linearities and uncertainties, leading to sub-optimal solutions.
    \item Optimization-based methods, on the other hand, formulate the VIO problem as an optimization task, where the goal is to minimize the overall reprojection error of visual features and the IMU residuals. These methods typically use bundle adjustment techniques to optimize the camera poses and landmarks jointly. Examples of optimization-based VIO algorithms are OKVIS \cite{leutenegger2013keyframe}, ORB-SLAM\cite{mur2015orb}, and ROVIO \cite{bloesch2015robust}. These algorithms often achieve higher accuracy compared to filter-based methods but require more computational resources.
\end{itemize}
\subsection{Adaptive Noise Estimation}
Accurate estimation of sensor noise is crucial for obtaining reliable state estimates in VIO algorithms. Traditional methods typically rely on precalibrated noise covariance matrices, obtained during sensor calibration. However, these constant noise models do not account for the dynamic nature of sensor noise, which can vary based on environmental conditions and sensor health.

To address this limitation, researchers have explored adaptive noise estimation techniques. These methods aim to adjust the noise covariance dynamically based on the observed sensor data, allowing the algorithm to adapt to changing noise characteristics. Some model-based approaches \cite{mehra1970identification},  \cite{zhang2020identification}, \cite{he2021adaptive} utilize statistical models or filtering techniques to estimate and update the noise covariance during operation. These techniques are primarily designed for filter-based algorithms and may not be directly applicable to optimization-based VIO methods.

\subsection{Deep Learning for SLAM and VIO}

Deep learning has shown significant promise in various aspects of SLAM and VIO. Convolutional neural networks (CNNs) \cite{o2015introduction} have been employed for feature detection and matching, leading to improved visual odometry \cite{mahjourian2018unsupervised} and loop closure detection \cite{chen2017deep}.
Different deep learning techniques have also been explored to estimate camera poses \cite{clark2017vinet}, predict depth maps \cite{godard2017unsupervised}, and improve feature tracking \cite{quach2021supslam}. These methods have demonstrated the potential to enhance VIO algorithm performance and robustness.
Other research focuses on improving the inertial performance using learning based methods. InertialNet \cite{liu2019inertialnet} inputs camera and IMU measurements to a CNN model that estimates the camera motion. RNIN-VIO \cite{chen2021rnin} uses a long short-term memory network (LSTM) \cite{yu2019review} to estimate the current position using IMU measurements and previous states from visual-inertial fusion. In \cite{buchanan2022deep} the authors adaptively estimate IMU bias for factor graph problems using LSTM and Transformers \cite{vaswani2017attention}. OriNet \cite{esfahani2019orinet} authors utilized a deep learning framework for a 3D orientation estimation using a single IMU.

Our proposed approach leverages deep learning for adaptive noise variance estimation, enabling VIO algorithms to dynamically estimate and update the IMU noise uncertainty, leading to improved accuracy and robustness.

\begin{figure*}[h!]
\begin{center}
\includegraphics[scale=0.162]{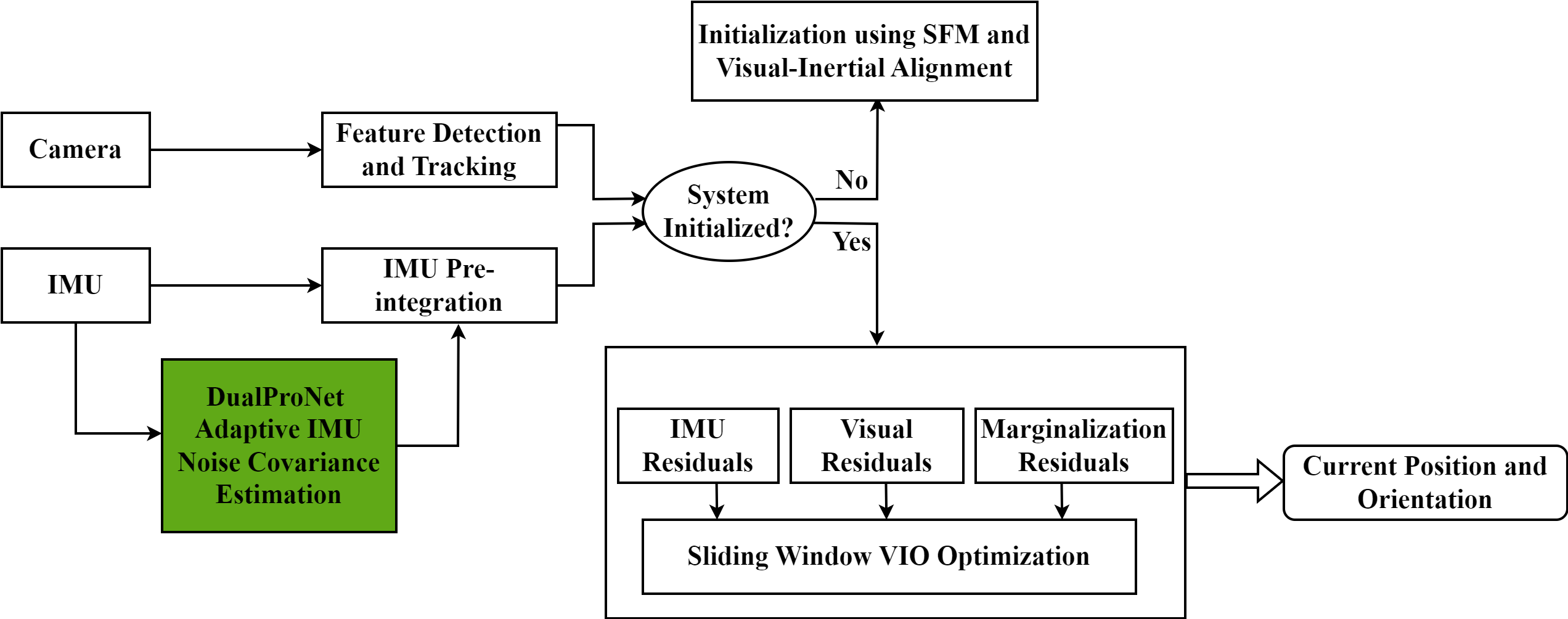}
\caption{VIO-DualProNet architecture. The baseline VIO algorithm is presented with our DualProNet addition colored in green.}
\label{fig:vio-dualpronet}
\end{center}
\end{figure*}
\section{Proposed Approach}\label{sec:PA}
In this section, we describe our proposed approach for visual-inertial odometry with learning-based adaptive noise covariance estimation. We begin by presenting an overview of the framework, followed by the details of our DualProNet deep neural network architecture and its integration into a VIO framework.\\
Our proposed approach aims to improve the accuracy and robustness of VIO algorithms by dynamically estimating the inertial sensor noise covariance using deep learning techniques. The framework consists of two main components: a deep neural network for noise covariance regression, DualProNet, and its integration with the VINS-Mono algorithm as described below:
\begin{enumerate}
    \item \textbf{DualProNet – A Deep Neural Network for Noise Covariance Estimation}: The deep neural network is responsible for estimating the inertial sensors noise covariance matrix based on sensor measurements. It takes as input the IMU sensor data, including specific force and angular velocity measurements, and outputs the current noise covariance. The network is trained using a dataset of sensor measurements and corresponding ground truth noise covariance values. DualProNet refers to our network architecture because of the different characteristics of accelerometers and gyroscopes. 
    \item \textbf{VIO-DualProNet}:  The estimated noise covariance from DualProNet is fed into the well established VINS-Mono algorithm as the dynamic noise model for the inertial sensors. During VIO optimization, the noise covariance is updated at each time step, allowing the algorithm to adapt to changing noise characteristics that often occur in real-world scenarios. The adaptive noise model aims to enhance the VIO algorithm's performance, particularly in challenging scenarios, dynamic environments, and during inertial sensor performance degradation.
\end{enumerate}

VINS-Mono \cite{qin2018vins} is an optimization based state-of-the-art VIO algorithm that has gained massive use and popularity over the last few years. It incorporates visual data from a monocular camera and inertial data from an IMU sensor. The optimization stage of the algorithm optimizes the current position and orientation of the system by solving a nonlinear factor graph optimization problem. In our study, we chose to intentionally disable the loop closure functionalities of VINS-Mono to amplify the influence of accurate noise covariance estimation on position and orientation estimation. Loop closure modules are advantageous in familiar or previously traversed areas as they help reduce accumulated drift. However, in numerous common scenarios where the trajectory is unexplored, a reliable VIO solution becomes indispensable for ensuring a high-quality navigation system. This compels the system to be more robust and accurate without relying on loop closure mechanisms. The VIO-DualProNet system architecture is shown in Fig. \ref{fig:vio-dualpronet} where our DualProNet block is colored in green. \\
The following section describes the main building blocks  of the VIO-DualProNet system architecture.
\subsection{VIO Factor Graph Optimization}
\begin{figure}[h!]
\begin{center}
\captionsetup{justification=centering}
\includegraphics[width= 8.9cm, height=4cm]{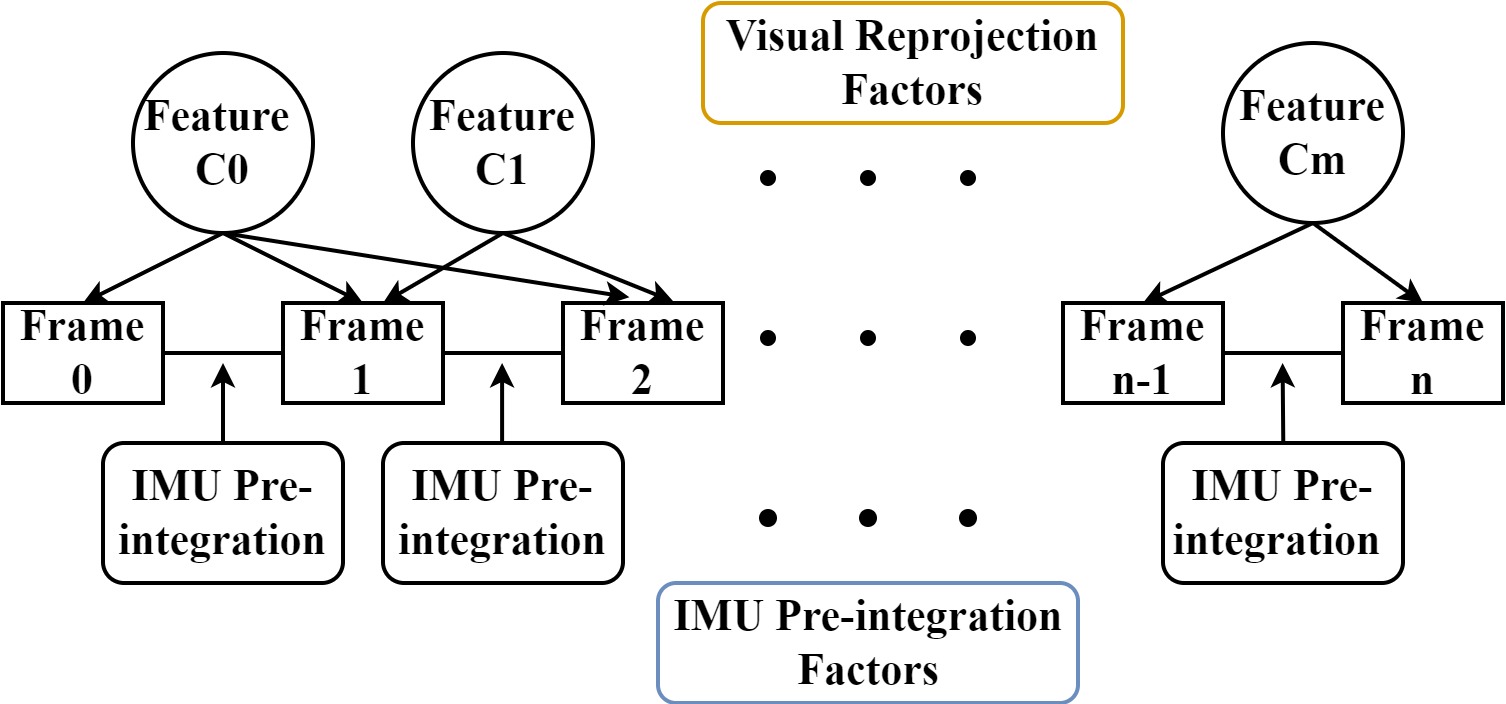}
\caption{The sliding window process in VIO algorithms, including IMU pre-integration factors between consecutive frames and visual features reprojection factors. } 
\label{fig:sliding-window}
\end{center}
\end{figure}

Optimization-based SLAM algorithms utilize a sliding window based on keyframes to construct a factor graph model. The model comprises various factor nodes, including visual reprojection factors, IMU pre-integration factors, and marginalization factors. Each factor corresponds to the residual of the error equation derived from the respective sensor's relative measurements. The marginalization factor acts as a prior constraint by fixing certain previous states during sliding window updates. It is obtained through the Schur complement operation \cite{zhang2006schur} involving the preceding sensor factor. The sliding window diagram is shown in Fig. \ref{fig:sliding-window}.\\
The state  \textbf{X} comprises the navigation states $\textbf{\textit{x}}$, extrinsic parameters $\textbf{\textit{x}}_c^b$, and visual features depths $\lambda$ is defined as:
\begin{equation}
\textbf{X} = [\textbf{\textit{x}}_0, \textbf{\textit{x}}_1, \ldots, \textbf{\textit{x}}_n, \textbf{\textit{x}}_c^b, \lambda_0, \lambda_1, \ldots, \lambda_m]
\end{equation}
where $\textit{n}$ is the number of keyframes, $m$ is the number of visual features in the sliding window, and $\lambda_l$ is the inverse depth of the $l\textsuperscript{th}$ feature from its first observation.\\
The navigation state at time (of frame) $k$ is formulated as
\begin{equation}
\textbf{\textit{x}}_k = [{\textbf{\textit{p}}_{b_k}^w}^T, {\textbf{\textit{q}}_{b_k}^w}^T, {\textbf{\textit{v}}_{b_k}^w}^T, \textbf{\textit{b}}_{f_b}^T, \textbf{\textit{b}}_{{\omega_b}}^T]^T, \quad k \in [0, n]
\end{equation}
where $\textbf{\textit{p}}_{b_k}^w$, $\textbf{\textit{q}}_{b_k}^w$, $\textbf{\textit{v}}_{b_k}^w$ are the position, orientation quaternion, and velocity expressed in the world frame at time $k$, and $\textbf{\textit{b}}_{f_b}$ and $\textbf{\textit{b}}_{{\omega_b}}$ are the bias terms of the specific force and the angular velocity expressed in the body frame, respectively. \\
The extrinsic rotation and translation matrices between the camera and the IMU body frame are represented as
\begin{equation}
\textbf{\textit{x}}_c^b = [{\textbf{\textit{p}}_c^b}^T, {\textbf{\textit{q}}_c^b}^T]^T
\end{equation}
where $\textbf{\textit{p}}_c^b$ is the translation and $\textbf{\textit{q}}_c^b$ is the rotation from the camera frame to the body frame.
\newline The visual-inertial, bundle adjustment optimization is formulated as a nonlinear, least-squares optimization problem:
\begin{equation}
\begin{split}
\min_\textbf{X} \Big\{ & \|\textbf{\textit{r}}_p - \textbf{H}_p \textbf{X}\|^2 + \sum_{k \in \textit{B}} \|\textbf{\textit{r}}_\textit{B} (\hat{\textbf{z}}_{{b}_k}^{b_{k+1}}, \textbf{X})\|_{\textbf{P}_{{b}_k}^{b_{k+1}}}^2 \\
&+ \sum_{(i,j) \in \textit{C}} \rho(\|\textbf{\textit{r}}_C (\hat{\textbf{z}}_{{l}}^{c_j}, \textbf{X})\|_{\textbf{P}_l^{c_j}}^2) \Big\}
\end{split} \label{eq:residuals}
\end{equation}
where $\textbf{\textit{r}}_p, \textbf{H}_p$ are the prior information from marginalization, $\textbf{\textit{r}}_B (\hat{\textbf{z}}_{{b}_k}^{b_{k+1}}, \textbf{X})$ is the IMU factor residual described in detail in Section~\ref{subsec:imu-factor}, the visual residual formulated as $\textbf{\textit{r}}_C (\hat{\textbf{z}}_{{l}}^{c_j}, \textbf{X})$ is described in Section~\ref{subsec:visual-residual}, and $\rho$ represents the Huber norm \cite{huber1992robust} used for regularization.
\vspace{\baselineskip}
\subsection{IMU Pre-integration Factor}
\label{subsec:imu-factor}
The IMU measurement model formulates the relation between the measured raw values and the real values while considering the noises and biases. Commonly, the underlying assumption is that the inertial sensor measurements have a constant variance, which is often not the case. Here, we incorporate our DualProNet regression to adaptively estimate the current variance of the inertial measurements.  \\
The measurement model for the measured specific force and angular velocity vectors is
\begin{equation}
\hat{\textbf{\textit{f}}}_{b_t} = \textbf{\textit{f}}_{b_t} + \textbf{R}_{\omega}^t \cdot \textbf{\textit{g}}^w + \textbf{\textit{b}}_{f_{b_t}} + \textbf{\textit{n}}_{f_b} 
\end{equation}
\begin{equation}
\hat{\bm{\omega}}_{b_t} = \bm{\omega}_{b_t} + \textbf{\textit{b}}_{{\omega}_{b_t}} + \textbf{\textit{n}}_{{\omega}_b}
\end{equation} 
where $\hat{\textbf{\textit{f}}}_{b_t}$ and $\hat{\bm{\omega}}_{b_t}$ are the measured specific force and angular velocity in the body frame at time $t$, respectively, $\textbf{\textit{f}}_{b_t}$ is the true specific force, $\bm{\omega}_{b_t}$ is the true angular velocity, $\textbf{R}_{\omega}^t$ stands for the rotation matrix between the body frame and the world frame, and $\textbf{\textit{g}}^w$ is the gravity force expressed in world frame. Considering the inertial error terms, $\textbf{\textit{b}}_{f_{b_t}}$ and $\textbf{\textit{b}}_{{\omega}_{b_t}}$ represent the specific force and angular velocity biases at time \textit{t} expressed in the body frame, respectively, and $\textbf{\textit{n}}_{f_b}$ and $\textbf{\textit{n}}_{{\omega}_b}$ describe the specific force and angular velocity noises modeled as additive Gaussian noises with zero mean:
\begin{equation}\label{eq:acc_noises}
\textbf{\textit{n}}_{f_b} \sim N(0, \bm{\sigma}_f^2)
\end{equation}
\begin{equation}\label{eq:gyr_noises}
\textbf{\textit{n}}_{{\omega}_b} \sim N(0, \bm{\sigma}_{\omega}^2)
\end{equation}
where $ \bm{\sigma}_f$ is the specific force noise standard deviation and $ \bm{\sigma}_{\omega}$ is the angular velocity noise standard deviation. As later explained, our proposed approach uses DualProNet to regress $ \bm{\sigma}_f$ and  $ \bm{\sigma}_{\omega}$ at each epoch and thereby produce an accurate estimate of the inertial uncertainty.\\ 
The sensor biases are modeled as a random walk with a Gaussian derivative:
\begin{equation}
\dot{\textbf{\textit{b}}}_{f} = \textbf{\textit{n}}_{b_f},\,\ \textbf{\textit{n}}_{b_f} \sim N(0, \bm{\sigma}_{b_f}^2) 
\end{equation}
\begin{equation}
\dot{\textbf{\textit{b}}}_{\omega_t} = \textbf{\textit{n}}_{b_{\omega}},\,\ \textbf{\textit{n}}_{b_{\omega}} \sim N(0, \bm{\sigma}_{b_{\omega}}^2)
\end{equation}
where $\textbf{\textit{n}}_{b_f}$ is the specific force bias zero mean white noise, $\textbf{\textit{n}}_{b_{\omega}}$ is the angular velocity bias zero mean white noise, $\bm{\sigma}_{b_f}$ is the specific force random walk noise standard deviation, and $\bm{\sigma}_{b_{\omega}}$ is the angular velocity random walk noise.

A common practice for optimization-based, visual-inertial SLAM algorithms is to use a pre-integration model \cite{forster2015imu} for the IMU measurements. This approach allows for integrating only the specific force and angular velocities between consecutive frames, regardless of the initial conditions of the position and velocity of the previous frames. This eliminates the need to repropagate IMU measurements after the starting conditions change, saving computational resources.\\
When the inertial equations of motion expressed in the world frame are transformed to the body frame the navigation solution with pre-integration coefficients is derived as follows:
\begin{equation}
\textbf{R}_w^{b_k} \textbf{\textit{p}}_{b_{k+1}}^w = \textbf{R}_w^{b_k} (\textbf{\textit{p}}_{b_k}^w + \textbf{\textit{v}}_{b_k}^w \Delta t_k - \frac{1}{2} \textbf{\textit{g}}^w \Delta t_k^2) + \bm{\alpha}_{b_{k+1}}^{b_k} 
\end{equation}
\begin{equation}
\textbf{R}_w^{b_k} \textbf{\textit{v}}_{b_{k+1}}^w = \textbf{R}_w^{b_k} (\textbf{\textit{v}}_{b_k}^w - \textbf{\textit{g}}^w \Delta t_k) + \bm{\beta}_{b_{k+1}}^{b_k}    
\end{equation}
\begin{equation}
\textbf{\textit{q}}_w^{b_k} \otimes \textbf{\textit{q}}_{b_k+1}^w = \bm{\gamma}_{b_{k+1}}^{b_k}
\end{equation}
where $\textbf{R}_w^{b_k}$ and $\textbf{\textit{q}}_w^{b_k}$ represent the rotation from world frame to body frame in rotation matrix and quaternion representations respectively. $\Delta t_k$ is the time interval between frame k and frame k-1.
\newline The pre-integrated parts that are independent of the starting conditions are formulated like this:
\begin{equation}
\bm{\alpha}_{b_{k+1}}^{b_k} = \int_{t \in [k,k+1]} \textbf{R}_t^{b_k} \left( \hat{\textbf{\textit{f}}}_{b_t} - \textbf{\textit{b}}_{f_{b_t}} \right)^2 \, dt 
\end{equation}
\begin{equation}
\bm{\beta}_{b_{k+1}}^{b_k} = \int_{t \in [k,k+1]} \textbf{R}_t^{b_k} \left( \hat{\textbf{\textit{f}}}_{b_t} - \textbf{\textit{b}}_{f_{b_t}} \right) \, dt
\end{equation}
\begin{equation}
\bm{\gamma}_{b_{k+1}}^{b_k} = \int_{t \in [k,k+1]} \frac{1}{2} \bm{\Omega} \left( \hat{\bm{\omega}}_{b_t} - \textbf{\textit{b}}_{{\omega}_{b_t}} \right) \bm{\gamma}_t^{b_k} \, dt
\end{equation}

Here, $\Omega$ is constructed as
\begin{equation}
\bm{\Omega}(\bm{\omega})=\left[ \begin{array}{cc}
-[\bm{\omega}]_{\times} & \bm{\omega} \\
-\bm{\omega}^T & 0
\end{array} \right]
\end{equation}

where $[\hspace{4pt}]_{\times}$ is the skew-matrix operator:
\begin{equation}
[\bm{\Omega}]_{\times}=\left[ \begin{array}{ccc}
0 & -\omega_{z} & \omega_{y} \\
\omega_{z} & 0 & -\omega_{x} \\
-\omega_{y} & \omega_{x} & 0 \\
\end{array} \right].
\end{equation}
Between every two consecutive frames, a new integration base is established by setting $\bm{\alpha}_{b_k}^{b_k} = \bm{\beta}_{b_k}^{b_k} = 0$ and $\bm{\gamma}_{b_k}^{b_k}$ is set to an identity quaternion. 
\newline Then, the propagation terms are as follows:
\begin{equation}
\begin{aligned}
\hat{\bm{\alpha}}_{{i+1}}^{b_k} &= \hat{\bm{\alpha}}_{{i}}^{b_k} + \hat{\bm{\beta}}_{{i}}^{b_k} \delta t + \frac{1}{2} \textbf{R}(\hat{\bm{\gamma}}_{{i}}^{b_k})(\hat{\textbf{\textit{f}}}_{b_i} - \textbf{\textit{b}}_{f_{b_i}}) \delta t^2 
\end{aligned}
\end{equation} 
\begin{equation}
\begin{aligned}
\hat{\bm{\beta}}_{{i+1}}^{b_k} &= \hat{\bm{\beta}}_{{i}}^{b_k} + \textbf{R}(\hat{\gamma}_{{i}}^{b_k})(\hat{\textbf{\textit{f}}}_{b_i} - \textbf{\textit{b}}_{f_{b_i}}) \delta t   
\end{aligned}
\end{equation} 
\begin{equation}
\begin{aligned}
\hat{\bm{\gamma}}_{{i+1}}^{b_k} &= \hat{\bm{\gamma}}_{{i}}^{b_k} \otimes \left[ \mathbf{1} \quad \frac{1}{2}(\hat{\bm{\omega}}_{b_i} - \textbf{\textit{b}}_{{\omega}_{b_i}}) \delta t \right].   
\end{aligned}
\end{equation}
Then, using error-state kinematics, the dynamic error model is derived:
\begin{equation}
\resizebox{0.9\linewidth}{!}{%
$\displaystyle
\begin{aligned}
\dot{\delta \textbf{\textit{x}}}_t &= \begin{bmatrix} \dot{\delta \bm{\alpha}}_t^{b_k} \\ \dot{\delta \bm{\beta}}_t^{b_k} \\ \dot{\delta \bm{\gamma}}_t^{b_k} \\ \dot{\delta \textbf{\textit{b}}}_{f_{b_t}} \\ \dot{\delta \textbf{\textit{b}}}_{{\omega}_{b_t}} \end{bmatrix}
= \begin{bmatrix} \textbf{0} & \mathbf{I} & \textbf{0} & \textbf{0} & \textbf{0} \\ \textbf{0} & \textbf{0} & -\textbf{R}_t^{b_k}( \hat{\textbf{\textit{f}}}_{b_t} - \textbf{\textit{b}}_{f_{b_t}})_{\times} & -\textbf{R}_t^{b_k} & \textbf{0} \\ \textbf{0} & \textbf{0} & -(\hat{\bm{\omega}}_{b_t} - \textbf{\textit{b}}_{{\omega}_{b_t}})_{\times} & \textbf{0} & -\mathbf{I} \\ \textbf{0} & \textbf{0} & \textbf{0} & \textbf{0} & \textbf{0} 
\\ \textbf{0} & \textbf{0} & \textbf{0} & \textbf{0} & \textbf{0}  \end{bmatrix} \begin{bmatrix} \delta \bm{\alpha}_t^{b_k} \\ \delta \bm{\beta}_t^{b_k} \\ \delta \bm{\gamma}_t^{b_k} \\ \delta \textbf{\textit{b}}_{f_{b_t}} \\ \delta \textbf{\textit{b}}_{{\omega}_{b_t}} \end{bmatrix} \\
&\quad + \begin{bmatrix} \textbf{0} & \textbf{0} & \textbf{0} & \textbf{0} \\ -\textbf{R}_t^{b_k} & \textbf{0} & \textbf{0} & \textbf{0} \\ \textbf{0} & -\mathbf{I} & \textbf{0} & \textbf{0} \\ \textbf{0} & \textbf{0} & \mathbf{I} & \textbf{0} \\ \textbf{0} & \textbf{0} & \textbf{0} & \mathbf{I} \end{bmatrix} \begin{bmatrix} \textbf{\textit{n}}_{f_b} \\ \textbf{\textit{n}}_{{\omega}_{b}} \\ \textbf{\textit{n}}_{b_f} \\ \textbf{\textit{n}}_{b_{\omega}} \end{bmatrix}
\end{aligned}
$}
\end{equation}
or, in a simplified manner:
\begin{equation}
\dot{\delta \textbf{\textit{x}}}_t = \textbf{F}_t \delta \textbf{\textit{x}}_t + \textbf{G}_t.
\end{equation}
The corresponding estimated error state covariance $\hat{\textbf{P}}_{b_{k+1}}^{b_k}$ can be recursively computed using a first-order covariance update with initial zero covariance:
\begin{equation}
\hat{\textbf{P}}_{t+\delta t}^{b_k} = (\textbf{I} + \textbf{F}_t \delta t) \hat{\textbf{P}}_t^{b_k} (\textbf{I} + \textbf{F}_t \delta t)^T + (\textbf{G}_t \delta t) \hat{\textbf{Q}} (\textbf{G}_t \delta t)^T
\end{equation}
where \(\delta t\) is the time interval between a pair of IMU measurements, and $\hat{\textbf{Q}}$ is the diagonal noise covariance matrix consisting of \(\text{diag}(\bm{\hat{\sigma}_f}^2, \bm{\hat{\sigma}_\omega}^2, \bm{\sigma}_{b_f}^2, \bm{\sigma}_{b_\omega}^2)\) in its diagonal. \\
\begin{figure*}[t!]
\begin{center}
\includegraphics[scale=0.7]{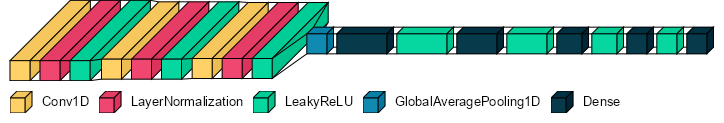}
\caption{Network architecture for regressing the inertial sensor measurement uncertainty.}
\label{fig:ProNet}
\end{center}
\end{figure*}
Note that we use DualProNet to adaptively  regress $\bm{\sigma}_f$ and  $\bm{\sigma}_\omega$ while $\bm{\sigma}_{b_f}$ and $ \bm{\sigma}_{b_\omega}$ are assumed to be constant. The motivation for this choice is that the latter standard deviations are a designer modeling selection that can be easily changed while the former standard deviations relates directly to the sensor hardware, which is sensitive to the operating environment and dynamics and thus changes throughout sensor operation. As a consequence the process matrix $\textbf{Q}$ and error state covariance $\textbf{P}$ are also denoted as estimated quantities.  
\newline The first-order Jacobian matrix needed for the optimization procedure can be computed as
\begin{equation}
\hat{\textbf{J}}_{t+\delta t} = (\textbf{I} + \textbf{F}_t \delta t) \hat{\textbf{J}}_t, \quad \hat{\textbf{J}}_{b_k}^{b_k} = \textbf{I}
\end{equation}
Therefore, the first order approximation of the pre-integration variables can be written like this:
\begin{equation}
\bm{\alpha}_{b_{k+1}}^{b_k} \approx \hat{\bm{\alpha}}_{b_{k+1}}^{b_k} + \hat{\textbf{J}}_{b_f}^\alpha \delta \textbf{\textit{b}}_{f_k} + \hat{\textbf{J}}_{b_\omega}^\alpha \delta \textbf{\textit{b}}_{\omega_k}
\end{equation}
\begin{equation}
\bm{\beta}_{b_{k+1}}^{b_k} \approx \hat{\bm{\beta}}_{b_{k+1}}^{b_k} + \hat{\textbf{J}}_{b_f}^\beta \delta \textbf{\textit{b}}_{f_k} + \hat{\textbf{J}}_{b_\omega}^\beta \delta \textbf{\textit{b}}_{\omega_k} 
\end{equation}
\begin{equation}
\bm{\gamma}_{b_{k+1}}^{b_k} \approx \hat{\bm{\gamma}}_{b_{k+1}}^{b_k} \otimes \begin{bmatrix} 1 \\ \frac{1}{2} \hat{\textbf{J}}_{b_\omega}^\gamma \delta \textbf{\textit{b}}_{\omega_k} \end{bmatrix}
\end{equation}
Thus, the inertial navigation state residual defined in \eqref{eq:residuals} can be written like this:
\begin{equation}
\resizebox{\linewidth}{!}{
$\textbf{\textit{r}}_\textit{B}(\hat{\textbf{z}}_{b_k}^{b_{k+1}}, \textbf{X}) = 
\begin{bmatrix}
\textbf{R}_w^{b_k} (\textbf{\textit{p}}_{b_{k+1}}^w - \textbf{\textit{p}}_{b_k}^w - \textbf{\textit{v}}_{b_k}^w \Delta t_k + \frac{1}{2} \textbf{\textit{g}}^w \Delta t_k^2) - \bm{\alpha}_{b_{k+1}}^{b_k} \\
\textbf{R}_w^{b_k} (\textbf{\textit{v}}_{b_{k+1}}^w - \textbf{\textit{v}}_{b_k}^w + \textbf{\textit{g}}^w \Delta t_k) - \bm{\beta}_{b_{k+1}}^{b_k} \\
2[\textbf{\textit{q}}_{b_k}^{w^{-1}} \otimes \textbf{\textit{q}}_{b_{k+1}}^w \otimes (\bm{\gamma}_{b_{k+1}}^{b_k})^{-1}]_{xyz} \\
\textbf{\textit{b}}_{a_{b_{k+1}}} - \textbf{\textit{b}}_{a_{b_k}} \\
\textbf{\textit{b}}_{\omega_{b_{k+1}}} - \textbf{\textit{b}}_{\omega_{b_k}}
\end{bmatrix}$
}
\end{equation}
\subsection{Visual Reprojection Factor} \label{subsec:visual-residual}
For the visual front-end, VINS-Mono extracts features from images using the Shi-Tomasi corner detector \cite{shi1994good}. Then, the previously detected features are tracked to the next frame using the Kanade–Lucas–Tomasi (KLT) optical flow algorithm \cite{lucas1981iterative}. An outlier rejection is performed using a fundamental matrix \cite{luong1996fundamental} with a random sample consensus (RANSAC) algorithm \cite{derpanis2010overview}. The features kept after the outlier rejection are undistorted and projected to a unit sphere. The reprojection error of each feature in the sliding window is formulated in the optimization process as

\begin{equation}
\textbf{\textit{r}}_\textit{C}(\hat{\textbf{z}}_l^{c_j}, \textbf{X}) = \begin{bmatrix} \textbf{\textit{b}}_1 & \textbf{\textit{b}}_2 \end{bmatrix}^T \left(\hat{\textbf{P}}_l^{c_j} - \frac{ \textbf{P}_l^{c_j}}{\|\textbf{P}_l^{c_j}\|}\right) 
\end{equation}
where $\textbf{\textit{b}}_1$ and $\textbf{\textit{b}}_2$ are the orthogonal basis that span the tangential  plane of $\hat{\textbf{P}}_l^{c_j}$ defined by
\begin{equation}
\hat{\textbf{P}}_l^{c_j} = \pi_c^{-1} \begin{bmatrix} \hat{u}_l^{c_j} \\ \hat{v}_l^{c_j} \end{bmatrix}
\end{equation}
where  $\pi_c^{-1}$ is the back projection from pixel to unit vector, $[u_l^{c_i} \quad v_l^{c_i}]$ is the first observation of feature $\textit{l}$, and $[\hat{u}_l^{c_j} \quad \hat{v}_l^{c_j}]$ is the observation of feature $\textit{l}$ in image $\textit{j}$. \\
The standard visual covariance in the tangent space is described by
\begin{equation}
\begin{split}
\textbf{P}_l^{c_j} = \textbf{R}_b^c \bigl(\textbf{R}_w^{b_j} \bigl\{\textbf{R}_{b_j}^w \bigl[\textbf{R}_c^b \bigl(\frac{1}{\lambda_l} \pi_c^{-1} \begin{bmatrix} u_l^{c_i} \\ v_l^{c_i} \end{bmatrix} \bigr) + \textbf{\textit{p}}_c^b \bigr] + \textbf{\textit{p}}_{b_i}^w - \textbf{\textit{p}}_{b_j}^w \bigr\} - \textbf{\textit{p}}_c^b \bigr)
\end{split}
\end{equation}
Finally, $\textbf{\textit{r}}_\textit{C}(\hat{\textbf{z}}_l^{c_j}, \textbf{X})$ is used in \eqref{eq:residuals} as the visual residual.
\subsection{DualProNet}
\begin{figure}[h]
\begin{center}
\captionsetup{justification=centering}
\includegraphics[scale=0.22]{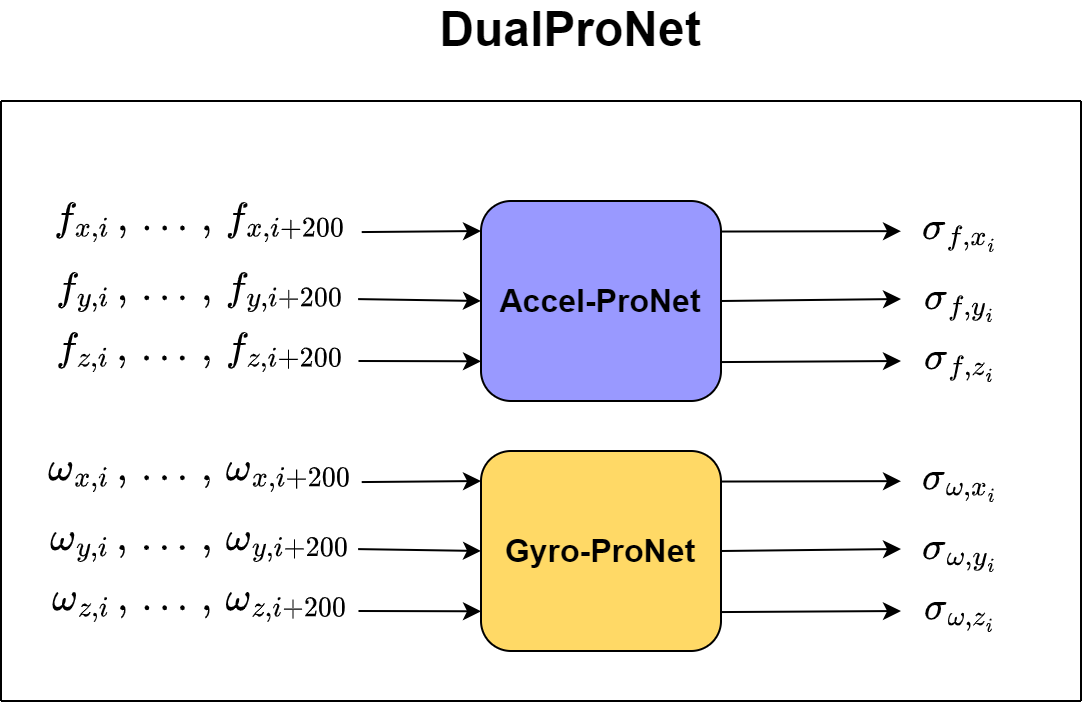}
\caption{Proposed DualProNet architecture, combining the Accel-ProNet model for regressing the accelerometer readings uncertainty and the Gyro-ProNet model for regressing the gyroscope readings uncertainty.}
\label{fig:DualProNet}
\end{center}
\end{figure}

A deep neural network architecture used for inertial process noise covariance estimation, ProNet, in a extended Kalman filter framework was  recently published in a set of papers \cite{or2022hybrid1, or2022hybrid, or2023pronet, or2023learning}.  The model architecture, presented in Fig. \ref{fig:ProNet}, consists of three one-dimensional convolutional layers, each followed by a leaky ReLU activation function \cite{xu2020reluplex} and layer normalization \cite{ba2016layer}. These convolutional layers play a crucial role in feature extraction from the input data, especially for sequential data like time series. By using learnable filters, they can capture important patterns and relationships present in the data. The subsequent application of the leaky ReLU activation function introduces non-linearity, enabling the model to learn more complex patterns. Layer normalization helps stabilize and accelerate the training process by normalizing the input along the feature dimension. After the convolutional layers, a global average pooling layer \cite{lin2013network} is applied, which computes the average value for each feature map, effectively summarizing the spatial information into a single value. This operation reduces the spatial dimensions of the data, aiding in reducing overfitting and computational costs.
The model concludes with four linear layers, each followed by a leaky ReLU activation function. These linear layers perform a linear transformation on the output from the previous layer, further enhancing the model's capacity to learn intricate relationships between features.
Together, this combination of layers allows the model to effectively extract relevant features, reduce spatial dimensions, and learn complex patterns, ultimately enabling it to make accurate predictions of the inertial noise covariance.  ProNet presents a novel approach to cope with varying process noise values, yet in field experiments it only slightly improved the existing model-based approaches. One of the reasons was the network structure taking as input both accelerometer and gyroscopes measurements together and assuming equal values in all sensor axes.\\
To circumvent this, we propose DualProNet, which is capable of regressing the noise parameters for each axis separately. Therefore, the model is constructed so that the input to the network is a consecutive sequence of measurements from a single axis of either the accelerometer or the gyroscope and the output of the network is the current estimated noise covariance of the input axis sequence, formulated as $\sigma_f$ and $\sigma_\omega$ in \eqref{eq:acc_noises}. Due to the different scale and characteristics of accelerometer readings compared to gyroscope readings, two different models were constructed and trained: Accel-ProNet for the specific force readings and Gyro-ProNet for the angular velocity readings. Our proposed DualProNet architecture is shown in Fig. \ref{fig:DualProNet}.
\section{Model Training and Fitting}\label{sec:MTF}
For the model training procedure, the dataset was first split into train and validation subsets and than the data was preprocessed to create proper training and validation sets for the covariance regression problem as described in this section.
\subsection{Dataset}
For network training and performance evaluation the commonly used Euroc-MAV \cite{burri2016euroc} dataset was chosen. This dataset consists of 11 recording sequences with an overall time of 22.5 minutes. The VIO sensors used in the dataset and in our research are MT9V034 global-shutter monochrome camera sampled at 20 Hz and an ADIS16448 MEMS IMU sampled at 200 Hz. For the ground truth generation, Euroc-MAV authors used a Leica Nova MS503 laser tracker and a Vicon motion capture system, providing position estimation with an accuracy down to the millimeter level.  
\newline The recorded sequences were divided into train and test subsets in the following manner; six sequences with a total of 12.5 minutes were the training set for the DualProNet training and five sequences with a total of 12 minutes were the test dataset. The trajectories were split randomly
to ensure that both datasets have scenarios with changing conditions. The full dataset split is shown in Table \ref{table:dataset}. 
\begin{table}[h]
\centering
\caption{Train-test split of the EuRoC MAV dataset.}
\label{table:dataset}
\begin{tabular}{|c|c|}
\hline
\textbf{Train (12.5m)}               & \textbf{Test (12m)}               \\
\hline
Machine Hall 01     & Machine Hall 02    \\
\hline
Machine Hall 03     & Machine Hall 04    \\
\hline
Machine Hall 05     & Vicon Room 1 01    \\
\hline
Vicon Room 1 02     & Vicon Room 1 03    \\
\hline
Vicon Room 2 01     & Vicon Room 2 02    \\
\hline
Vicon Room 2 03     &                    \\
\hline
\end{tabular}
\end{table}
\subsection{Data Preprocessing}
First, we created a ground-truth noise variance values for each subset of measurements. Thus, the measurements of each axis were divided to subsets of 200x1 length, as the networks' input size. Than, the measurements were filtered to remove noise using Savitzky-Golay filter \cite{press1990savitzky}. To train the model, Gaussian noise was added to the IMU samples with different variance values for the specific force and the angular velocity:
\begin{center}
$q_{\text{f}}$: [0.01, 0.03, 0.05, 0.07, 0.09, 0.11, 0.13, 0.15, 0.17, 0.19, 0.21] m/s\textsuperscript{2} \\
$q_{\omega}$: [0.001, 0.003, 0.005, 0.007, 0.009, 0.011, 0.013, 0.015] rad/s
\end{center}
The noise was uniformly introduced to all axes of a single sensor, whether an accelerometer or a gyroscope. This approach aimed to train the model to recognize and adapt to various magnitudes of noise across all axes, enhancing the model's resilience to diverse types of axis-related noise.
\subsection{Model Training}\label{sec:MT}
Our Accel-ProNet and Gyro-ProNet models were trained as a regression problem where the loss function was the mean square error (MSE) between the estimated values and the added noise covariances:
\begin{equation}
\textbf{MSE} = \frac{1}{n} \sum_{i=1}^{n} (\hat{Y}_i - Y_i)^2
\end{equation}
where $\textit{n}$ is the number of sequences in the data, $\hat{Y}_i$ is the estimated noise covariance value output by the network corresponding to the \textit{i}\textsuperscript{\textit{th}} sequence, and $Y_i$ is the true noise covariance of the \textit{i}\textsuperscript{\textit{th}} sequence.
\newline The training process of the models was implemented in a conda environment, utilizing the ADAM optimizer \cite{kingma2014adam} for the optimization process. The training was performed over 200 epochs, with a fixed learning rate of 0.001 and a batch size of 200. This choice of optimization algorithm and hyperparameters allowed the models to efficiently update their parameters during training, enabling them to converge to optimal noise covariances.

\begin{table*}[h]
\centering
\caption{VINS-Mono ATE [m] performance on Euroc-MAV dataset using different constant noise covariances against our method.}
\resizebox{2\columnwidth}{!}{%
\renewcommand{\arraystretch}{1.4}
\label{table:vio-results}
\begin{tabular}{|c|*{5}{c|}}
\hline
\multirow{2}{*}{\text{Sequence}} 
& \multicolumn{1}{c|}{$\sigma_{\textit{f}}$= 0.04 m/s\textsuperscript{2}, $\sigma_{\omega}$= 0.002 rad/s} 
& \multicolumn{1}{c|}{$\sigma_{\textit{f}}$= 0.08 m/s\textsuperscript{2}, $\sigma_{\omega}$= 0.004 rad/s} 
& \multicolumn{1}{c|}{$\sigma_{\textit{f}}$= 0.16 m/s\textsuperscript{2}, $\sigma_{\omega}$= 0.008 rad/s} 
& \multicolumn{1}{c|}{\text{VIO-DualProNet (ours)}} 
& \multicolumn{1}{c|}{\text{Rate of Improvement}} \\
& (Division by 2) & (Original baseline values) & (Multiplication by 2) & & (compared to baseline) \\
\hline
 MH 01 & 0.21 & \textbf{0.20} & 0.26 & 0.22 & -10.0\% \\
\hline
MH 02 & 0.18 & \textbf{0.17} & 0.20 & 0.19 & -11.7\% \\
\hline
MH 03 & 0.37 & 0.22 & 0.12 & \textbf{0.11} & 50.0\% \\
\hline
MH 04 & 0.48 & 0.38 & 0.35 & \textbf{0.30} & 21.0\% \\
\hline
MH 05 & 0.40 & 0.35 & 0.33 & \textbf{0.29} & 17.1\% \\
\hline
V1 01 & 0.11 & 0.08 & 0.06 & \textbf{0.05} & 37.5\% \\
\hline
V1 02 & 0.12 & 0.10 & 0.09 & \textbf{0.08} & 20.0\% \\
\hline
V1 03 & 0.32 & 0.15 & 0.13 & \textbf{0.11} & 26.6\% \\
\hline
V2 01 & 0.12 & 0.09 & 0.06 & \textbf{0.05} & 44.4\% \\
\hline
V2 02 & 0.26 & 0.13 & 0.08 & \textbf{0.07} & 46.1\% \\
\hline
V2 03 & 0.54 & 0.29 & 0.20 & \textbf{0.18} & 37.9\% \\
\hline
\end{tabular}
}
\end{table*}
\section{Experimental Results}\label{sec:ER}
This section describes the experimental results and performance evaluation of our DualProNet approach. We first present the performance of the deep neural networks on the generated train and test datasets (open loop), and then compare the adaptive noise covariance estimation using our deep learning models to the commonly used, model-based, constant noise models using the VINS-Mono algorithm (closed loop). The system was tested using Intel i7-1165 CPU at 2.8GHz.
\subsection{DualProNet Evaluation – Open Loop}
The regression of the inertial uncertainty is examined in an open-loop fashion, and thus does not depend on the VIO algorithm. The model performance on the train and test datasets was evaluated using root mean squared error (RMSE) criteria:
\begin{equation}
\textbf{RMSE} = \sqrt{\frac{\sum_{i=1}^{n} (\hat{X}_i - X_i)^2}{n}}
\end{equation}
where $\textit{n}$ is the number of sequences in the data, $\hat{X}_i$ is the estimated noise covariance value output by the network corresponding to the \textit{i}\textsuperscript{\textit{th}} sequence, and $X_i$ is the true noise covariance of the \textit{i}\textsuperscript{\textit{th}} sequence.
\newline Model training information is described in Section~\ref{sec:MT}. The DualProNet performance on the specific force and angular velocity uncertainties are shown in Table \ref{table:networks-rmse}. The specific force row represents the Accel-ProNet performance and the angular velocity represents the Gyro-ProNet performance. The networks demonstrated low RMSE results both for the train and the test sequences, proving their effectiveness in noise covariance estimation.

\begin{table}[h!]
\centering
\caption{RMSE of the inertial uncertainty using DualProNet.}
\renewcommand{\arraystretch}{1.5}
\label{table:networks-rmse}
\begin{tabular}{|c|c|c|}
\hline
 & Train RMSE & Test RMSE \\
\hline
Specific force [m/s$^2$] & 0.0253 & 0.0301 \\
\hline
Angular velocity [rad/s] & 0.00162 & 0.00185 \\
\hline
\end{tabular}
\end{table}

\subsection{VIO-DualProNet Evaluation – Closed Loop}
\subsubsection{Performance Metric}
The absolute trajectory error (ATE) metric is employed to evaluate the baseline and DualProNet VIO algorithm's accuracy and robustness. The ATE measures the difference between the estimated trajectory and the ground truth trajectory in 3D space. 
The quantitative formulation of ATE is as follows:
\begin{equation}
 \textbf{ATE} = \sqrt{\frac{1}{n} \sum_{i=1}^{n}  \| \textbf{\textit{p}}_{gt,i} - \textbf{\textit{p}}_{est,i} \|^2 } 
\end{equation}
where $\textit{n}$ is the number of time samples in the sequence, $\textbf{\textit{p}}_{gt,i}$ is the ground truth position at time i, and $\textbf{\textit{p}}_{est,i}$ is the system's estimated position at the \textit{i}\textsuperscript{\textit{th}} time sample.
\subsubsection{VIO Baselines}
To evaluate our approach against  constant covariance VIO approaches, we utilized the VINS-Mono algorithm and evaluated its performance on the Euro-MAV dataset. We examined three different sets of constant noise covariances. The first set is the original noise covariances used in the VINS-Mono algorithm (baseline), and the other sets comprise division and multiplication of the original values by a factor of two, a common practice in VIO algorithms. The full covariance tested sets:
\begin{enumerate}
  \item $\sigma_f= 0.04 m/s^2$, $\sigma_\omega$= 0.002 rad/s
    \item $\sigma_f= 0.08 m/s^2$, $\sigma_\omega$= 0.004 rad/s (original values baseline)
    \item $\sigma_f= 0.16 m/s^2$, $\sigma_\omega$= 0.008 rad/s
\end{enumerate}
The trained models were integrated into VINS-Mono and their localization accuracy was also tested. 
\begin{figure*}[h!]
    \centering
    
    \begin{tabular}{cccc}
    \hspace*{-1.6cm}
        \begin{subfigure}{0.28\textwidth}
            \includegraphics[width=\linewidth]{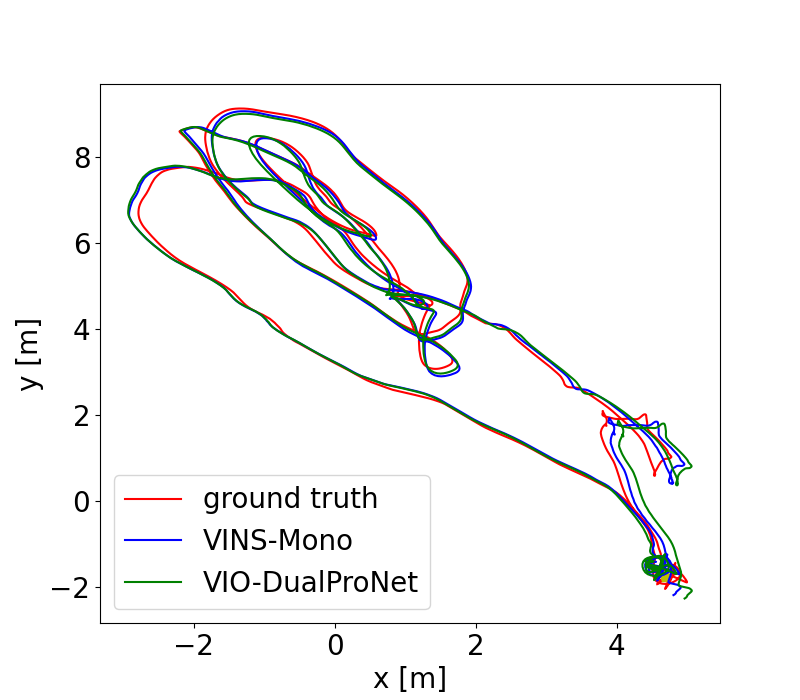}
            \caption{MH 01}
        \end{subfigure}
        \hfill
        \begin{subfigure}{0.28\textwidth}
            \includegraphics[width=\linewidth]{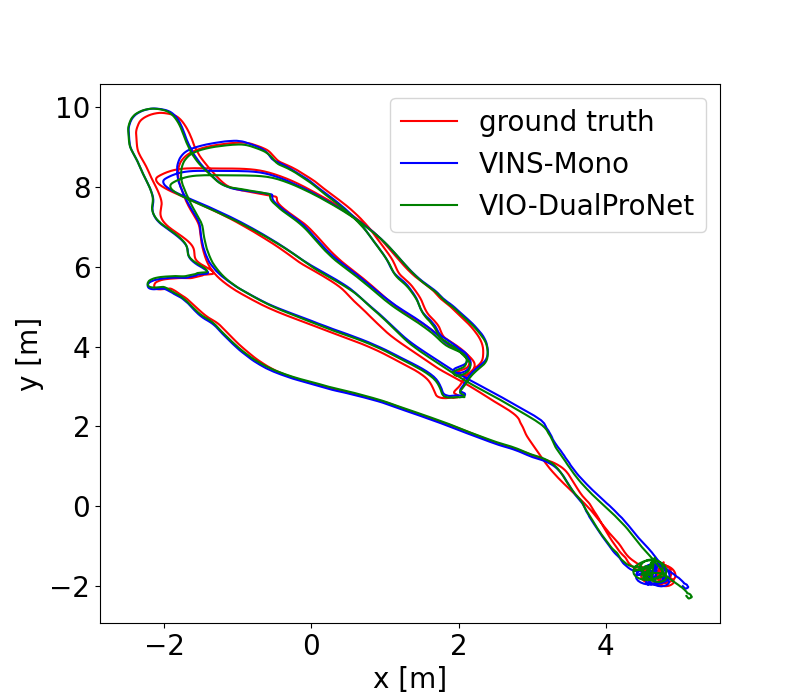}
            \caption{MH 02}
        \end{subfigure}
        \hfill
        \begin{subfigure}{0.28\textwidth}
            \includegraphics[width=\linewidth]{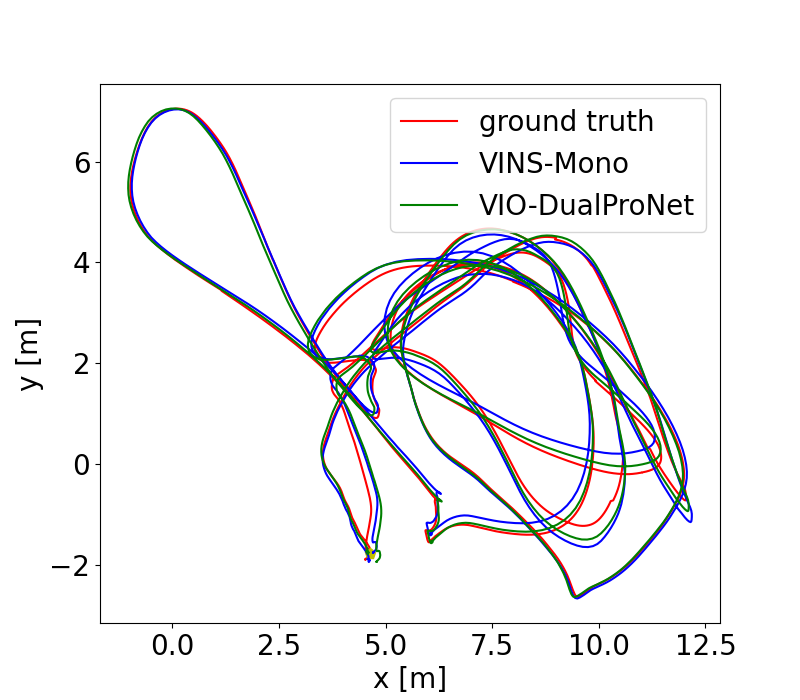}
            \caption{MH 03}
        \end{subfigure}
        \hfill
        \begin{subfigure}{0.28\linewidth}
            \includegraphics[width=\linewidth]{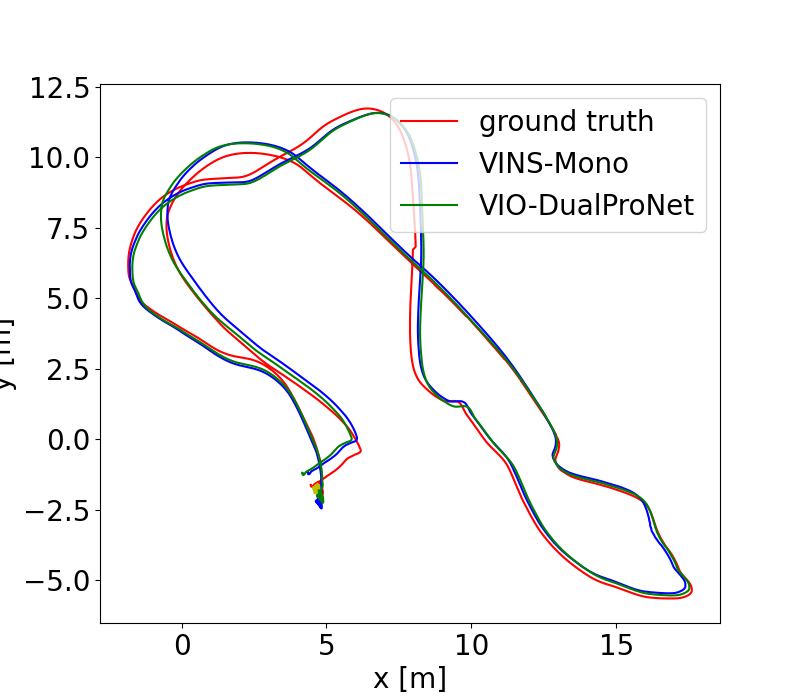}
            \caption{MH 04}
        \end{subfigure}
        
        \\
        \hspace*{-1.6cm}
        \begin{subfigure}{0.28\linewidth}
            \includegraphics[width=\linewidth]{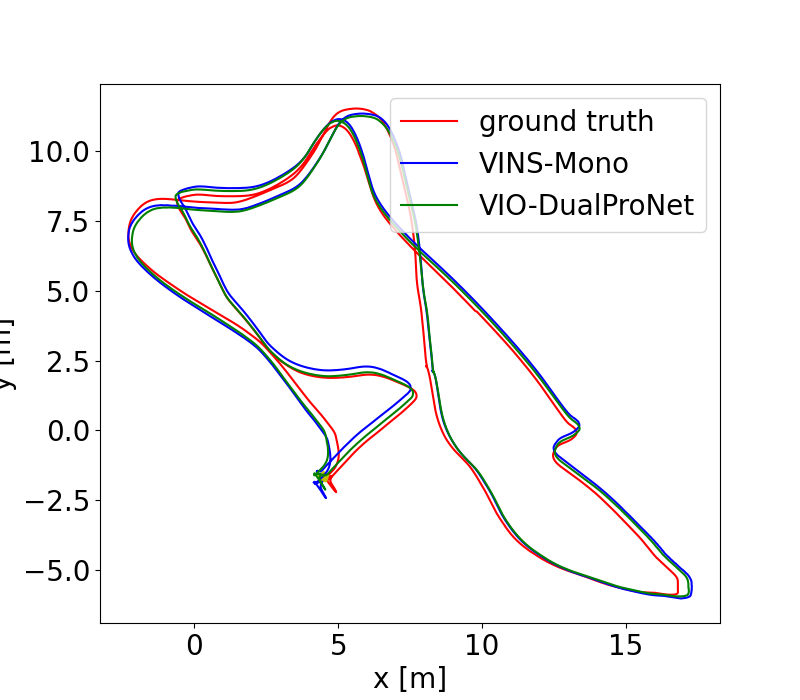}
            \caption{MH 05}
        \end{subfigure}
        \hfill
        \begin{subfigure}{0.28\linewidth}
            \includegraphics[width=\linewidth]{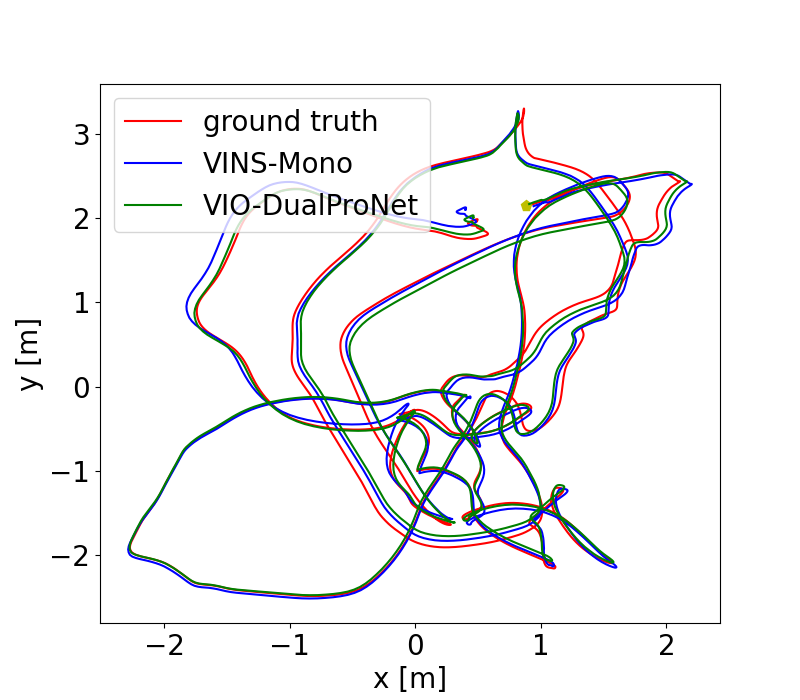}
            \caption{V1 01}
        \end{subfigure}
        \hfill
        \begin{subfigure}{0.28\linewidth}
            \includegraphics[width=\linewidth]{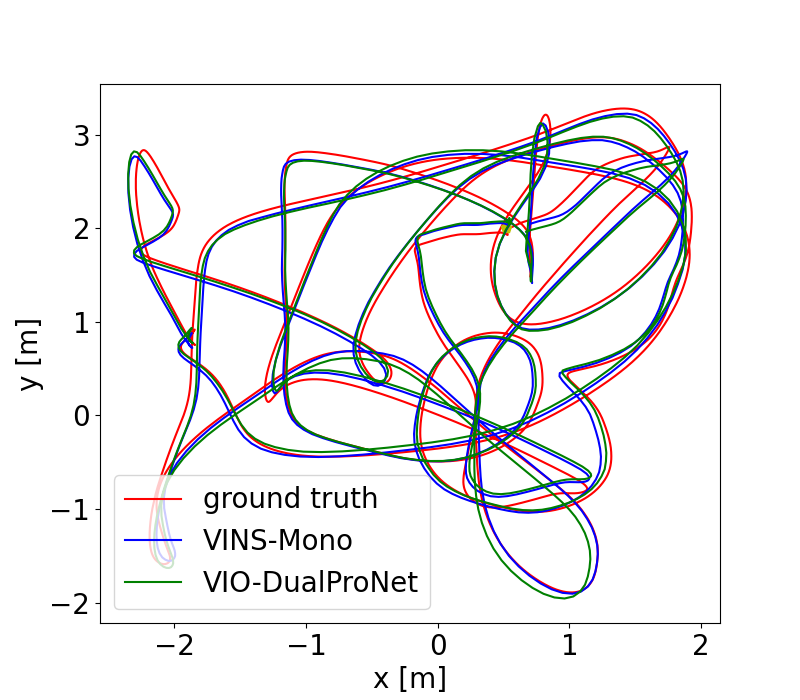}
            \caption{V1 02}
        \end{subfigure}
        \hfill
        \begin{subfigure}{0.28\linewidth}
            \includegraphics[width=\linewidth]{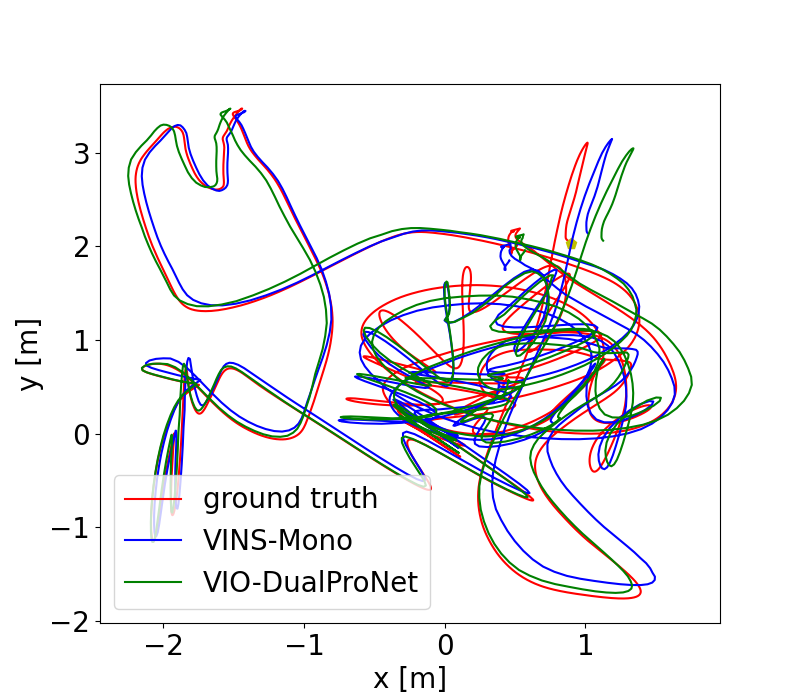}
            \caption{V1 03}
        \end{subfigure}
        
        \\
        \begin{subfigure}{0.28\linewidth}
            \includegraphics[width=\linewidth]{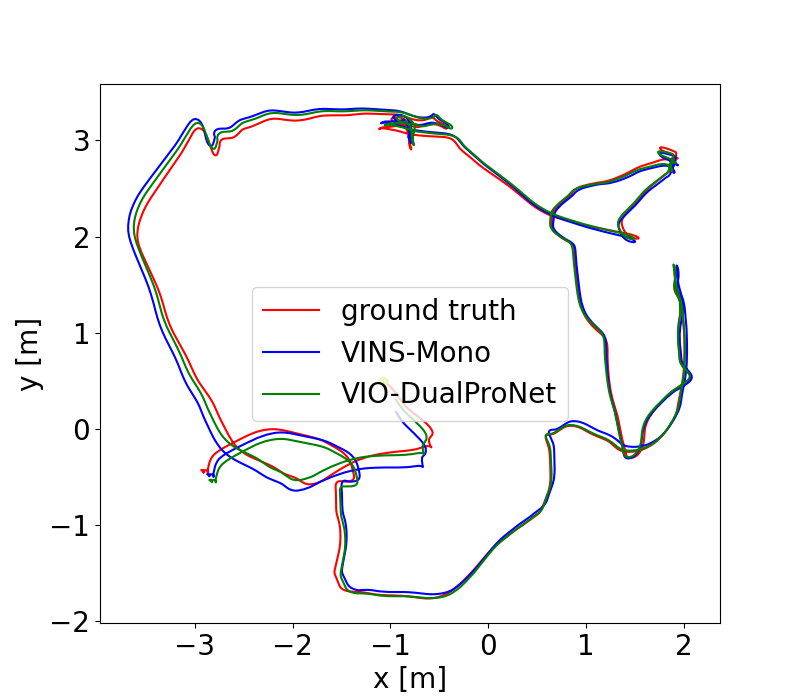}
            \caption{V2 01}
        \end{subfigure}
                
        \begin{subfigure}{0.28\linewidth}
            \includegraphics[width=\linewidth]{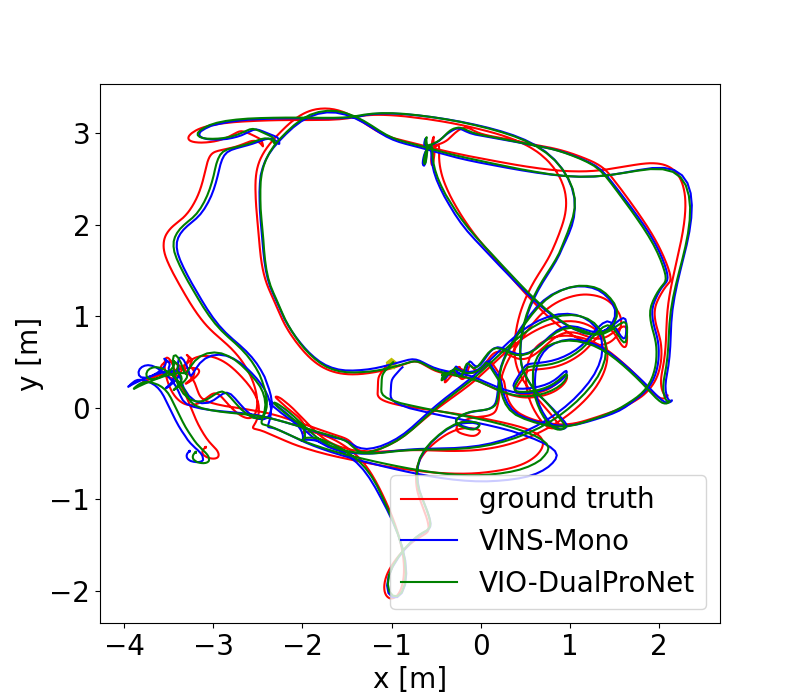}
            \caption{V2 02}
        \end{subfigure}
        
        \begin{subfigure}{0.28\linewidth}
            \includegraphics[width=\linewidth]{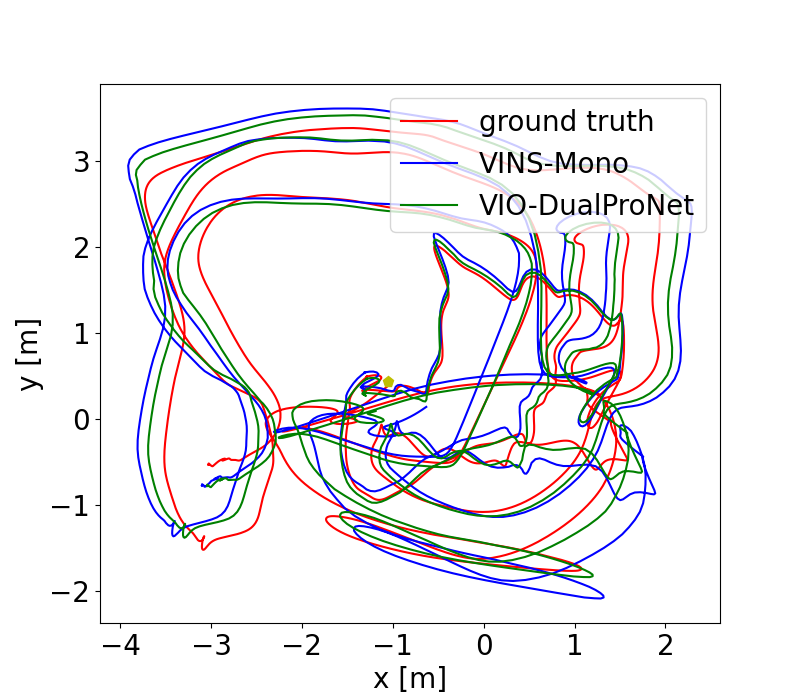}
            \caption{V2 03}
        \end{subfigure}
       
    \end{tabular}
    \captionsetup{justification=centering} 
    \caption{Euroc-MAV trajectories including the ground truth trajectories and the estimated trajectories from VINS-Mono using the original constant noise covariance values compared to our proposed VIO-DualProNet.}
    \label{fig:trajectories}
\end{figure*}
\subsubsection{Results}
The model-based VIO and our DualProNet-VIO performance is summarized in Table~\ref{table:vio-results}. From the results it can be seen that DualProNet-VIO outperformed the model-based constant covariance methods in 9 out of the 11 sequences, having a lower ATE than all other three tested constant covariance combinations. Our method achieved an average improvement of 25\% compared to the baseline approach. 
We also achieved an improvement of 12.5\% compared to the other best constant covariance combination, which is the multiplication by two combination. In some of the sequences our method achieved a great improvement, like sequences MH 03 and V2 02, reaching an improvement of 50\% and 46\% compared to baseline, respectively.\\
Acknowledging our overall positive outcomes, it is important to note that in sequences MH 01 and MH 02, our method exhibited slightly inferior results in comparison to the constant covariance approach. Despite the absence of discernible variations in IMU behavior in these sequences, we suspect that differences in visual quality might account for this outcome. Notably, accurate estimation of visual and inertial noise covariances significantly impacts visual-inertial fusion. However, our focus has predominantly centered on adaptive inertial noise covariance refinement, with the visual noise covariance remaining constant. This could potentially lead to less precise outcomes. These findings serve as motivation to further enhance the noise covariance estimation system to also include an adaptive visual quality covariance estimation module.
The entire trajectories are shown fully in Fig. \ref{fig:trajectories}. Each trajectory image is combined from the ground truth trajectory, the estimated trajectory from the VINS-Mono baseline algorithm, and the trajectory estimated by our VIO-DualProNet. As also shown in Table \ref{table:vio-results}, the trajectories of the VIO-DualProNet aligned better with the ground truth trajectories in the majority of the sequences, compared to the original VINS-Mono.
\section{Conclusion}\label{sec:CONC}
VIO algorithms are commonly used in robotics and autonomous platforms. One of its underlying assumptions is that the inertial uncertainly, in terms of its noise variance, is constant. As this assumption does not hold in real-world applications, we proposed DualProNet, a novel approach for visual-inertial odometry with learning-based adaptive noise covariance estimation. By dynamically estimating the inertial noise uncertainly using a dedicated deep neural network, we improve the accuracy and robustness of the VIO algorithm, particularly in challenging scenarios and dynamic environments. \\
The experimental results demonstrate the effectiveness of our proposed approach, with significantly improved accuracy and robustness compared to the baseline by an ATE improvement of $25\%$. The adaptive noise model allows the VIO algorithm to adapt to changing inertial noise characteristics, resulting in more reliable pose estimates. \\
Our approach opens up new opportunities for utilizing deep learning techniques in VIO applications, where adaptive noise estimation can play a crucial role in enhancing performance and robustness.
The advantage lies in eliminating the manual adjustment of constant IMU noise covariances, as our method automatically adapts to the specific noise levels present in each scenario. This adaptability not only simplifies the deployment of the VIO system in diverse environments but also ensures optimal performance and accuracy without the need for time-consuming parameter tuning. \\
In addition to VIO, DualProNet architecture can be applied to any optimization or estimation problems involving inertial sensors.

\bibliographystyle{IEEEtran}
\bibliography{ref}

\end{document}